# Tele-operative Robotic Lung Ultrasound Scanning Platform for Triage of COVID-19 Patients

Ryosuke Tsumura[†], *IEEE Member*, John W. Hardin[†], Keshav Bimbraw, Olushola S. Odusanya, Yihao Zheng, Jeffrey C. Hill, Beatrice Hoffmann, Winston Soboyejo, Haichong K. Zhang, *IEEE Member*

*Abstract*— Novel severe acute respiratory syndrome coronavirus 2 (SARS-CoV-2) has become a pandemic of epic proportions and a global response to prepare health systems worldwide is of utmost importance. In addition to its cost effectiveness in a resources-limited setting, lung ultrasound (LUS) has emerged as a rapid noninvasive imaging tool for the diagnosis of COVID-19 infected patients. Concerns surrounding LUS include the disparity of infected patients and healthcare providers, relatively small number of physicians and sonographers capable of performing LUS, and most importantly, the requirement for substantial physical contact between the patient and operator, increasing the risk of transmission. Mitigation of the spread of the virus is of paramount importance. A 2-dimensional (2D) tele-operative robotic platform capable of performing LUS in for COVID-19 infected patients may be of significant benefit. The authors address the aforementioned issues surrounding the use of LUS in the application of COVID-19 infected patients. In addition, first time application, feasibility and safety were validated in three healthy subjects, along with 2D image optimization and comparison for overall accuracy. Preliminary results demonstrate that the proposed platform allows for successful acquisition and application of LUS in humans.

## I. INTRODUCTION

Novel severe acute respiratory syndrome coronavirus 2 (SARS-CoV-2) has already become a pandemic of epic proportions, affecting over 36 million humans worldwide as of October 2020 [1]. Respiratory symptoms are the primary manifestation of COVID-19. The disease caused by SARS-CoV-2 can range from mild illness to severe, acute and fulminant respiratory distress. This varying severity in the face of a worldwide pandemic necessitates rapid diagnosis to provide the proper triage and disposition of patients. Diagnostic testing such as plain-film radiography (x-ray) and chest computed tomography (CT) are considered the gold-standard of diagnostic imaging in the detection of lung-related disease [2], [3]. However, in resource-limited areas, low-and middle-income countries, these imaging modalities are cost-prohibitive and not available at most of healthcare facilities [4].

Lung ultrasound (LUS) has emerged as an alternative for the rapid point-of-case diagnosis of pulmonary manifestations of COVID-19 with its advantage of low cost, safety, absence of radiation, and portability [3], [5], [6]. It has been widely adopted to image COVID-19 and clinical guidelines have been established. [7]–[9]. However, the diagnostic accuracy of LUS is highly operator-dependent and can be limited by patient habitus. In resource-limited areas, the accessibility of LUS is further restricted by the small number of emergency physicians and sonographers capable of performing LUS to accurately diagnose and triage COVID-19 patients. Additionally, conventional 2D ultrasound (US) exams require substantial physical contact between the operator and patient, increasing the risk of COVID-19 transmission.

### A. Related Works

We believe that robot-assisted US can play a critical role in addressing the aforementioned issues and potential limitations, because the robotic assistance enables tele-operative control of the US probe, making the procedure repeatable, and decreasing the interaction and contact between the user and patient [10], [11]. Since the late 1990s, the robotic US scan system has been investigated in various ways including tele-operative control [12]–[14], operator-robot corporative control [15], [16], and autonomous control [17]–[21]. Although most of this research proposed the robotic systems for general clinical applications, several studies investigated the system optimized for specific clinical applications such as fetal ultrasonography [22]–[26], focused assessment with sonography for trauma (FAST) [27], thyroid disease assessment [18], [28], [29], and echocardiography [30]. One recent study reported the successful cardiopulmonary assessment of patients with COVID-19 using a robot-assisted remote US system [31]. Meanwhile, to the best of our knowledge, there is no study to date that has developed a robotic US platform optimized for the application of LUS.

The robotic scan system for LUS is required to cover a large area of the thorax including the anterior, lateral, and posterior chests, and several standardized approaches have been developed for point-of-care LUS. For instance, the bedside LUS in emergency (BLUE)-protocol [5] is a widely disseminated and validated protocol as an adjunctive imaging approach for the immediate diagnosis of acute respiratory failure. Physicians using this protocol usually acquire US images to identify hallmark signs (e.g. pleural line, A-line, B-

[†] Equal contribution
Research supported by the National Institute of Health (DP5 OD028162).

R. Tsumura, K. Bimbraw and H. Zhang are with the Department of Biomedical Engineering, Worcester Polytechnic Institute, Worcester, MA 01609 USA (e-mail: hzhang10@wpi.edu).

J.C. Hill is with the Department of Diagnostic Medical Sonography, School of Medical Imaging and Therapeutics, MCPHS University, Worcester, MA 01608 USA

W. Soboyejo and Y. Zheng are with the Department of Mechanical Engineering, Worcester Polytechnic Institute, Worcester, MA 01609 USA

J. Hardin and B. Hoffmann are with the Department of Emergency Medicine, Beth Israel Deaconess Medical Center, Boston, MA 02215 USA

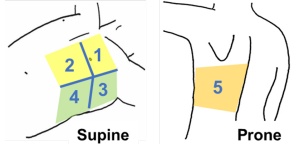

**Figure 1.** Standardized scan regions in the modified BLUE protocol. (anterior region: #1 and #2, lateral region: #3 and #4, posterior region: #5). Note that this illustration is cited as originally from [40].

line, and lung sliding) at ten standardized regions on the anterior, lateral, and posterior chest as shown in **Figure 1**. In addition, to acquire the diagnostic US images at each region, the probe positioning needs to be adjusted perpendicular with respect to the body surface.

To perform LUS under this protocol with conventional robotic approaches, multi-link kinematics is necessary and the patient safety, such as the avoidance of excessive contact force during the scan, relies on high precision sensors, a robust control algorithm, and computational performance. The acceptable contact force between the US probe and body may be less than approximately 20 N according to previous studies [23], [31]–[35], and needs to be maintained during the scan procedure regardless of the scan position. Additionally, it is unclear if the multi-link robot arm can cover all the scan regions required in the BLUE protocol that is timely and cost effective. One study noted that the robot US system could not reach some scan regions due to the robot configuration restriction [31]. In order to reach both sides of the chest with the robot arm while avoiding collisions between the robotic joints and the patient's body, an arm reach of at least 1200 mm is required considering the statistical data of the human body (see Section II-B). These requirements could be satisfied by a multi-link robotic configuration, which is cost prohibitive in a resource-limited setting.

*B. Contribution*

Our goal is to develop a tele-operative robotic LUS platform for COVID-19 diagnosis with a minimized risk of transmission between patients and healthcare workers in a resource-limited environment by significantly reducing physical contact during the procedure. To the best of our knowledge, this is the first robot designed for LUS. The limitations of prior studies suggest that the system configuration is not optimized for LUS in terms of the operational safety and probe reachability to all required chest areas. This manuscript reports an innovative approach that builds on a simple configuration with low-cost components, high usability with optimized kinematics, and safety features without relying on high precision sensors. We designed a first-of-its-kind prototype of a robotic LUS scan platform based on previous concepts and demonstrated its application in a pilot human subject study.

## II. MATERIALS AND METHODS

*A. System Overview*

**Figure 2 (a)** presents our proposed robotic LUS scan platform. This system allows the operator to tele-operatively manipulate the US probe based on the visual information captured by the cameras. Thus, the operator is not required to be present with the patient during the US procedure. The robotic platform is mainly composed of a gantry-style positioning unit and a passive-actuated end-effector. The positioning unit has an optimized number of joints and links, enabled to perform LUS based on current clinical guidelines: including the standardized 10-area LUS using the modified BLUE protocol [5]. Compared with a multi-link robotic arm, the proposed gantry-style kinematics provides the reachability and maneuverability to scan both hemithoraces with minimal risk of joint collision to the patient body. The gantry-style system is structurally simple and less costly to fabricate compared to the commercially available robotic arm and easy to implement in the resource-limited environment. In addition, the end-effector equips an electronics-free spring-based safety mechanism that maintains a constant contact force and a normal orientation with respect to the patient's skin surface. The position and posture of the US probe can be adjusted adaptively corresponding to the chest surface during the scanning procedure without any sophisticated control and computation. The electronics-free passive mechanical configuration can prevent the probe from applying excessive force and ensure patient safety. The robotic platform has a total of 8 axes: 4 axes at the gantry-style positioning unit and 4 axes at the passive-actuated end-effector. With the configuration of the robotic platform, we implemented an active-passive hybrid control to couple translation and orientation motions for a simple and intuitive manipulation interface. The manipulation of US probe is performed via a joy-stick operation. Also, three cameras covering top and two side views providing visual feedback for tele-operation. Importantly, the operator can communicate with the patient via a two-way microphone in real-time.

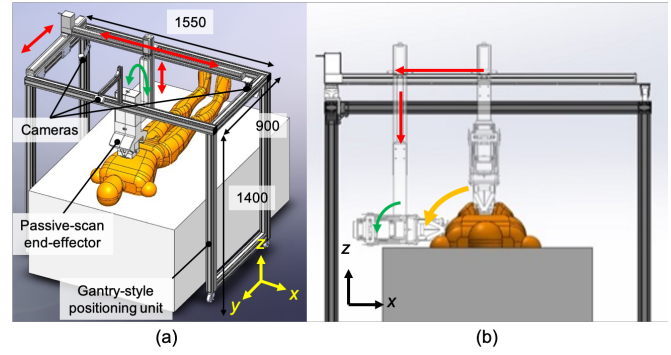

**Figure 2.** (a) Overview of robotic LUS scan platform comprised with a gantry-style positioning unit and passive-actuated end-effector (units: mm) (b) Scan on the side of chest by combining x- and z-axis translations and orientation of the end-effector.

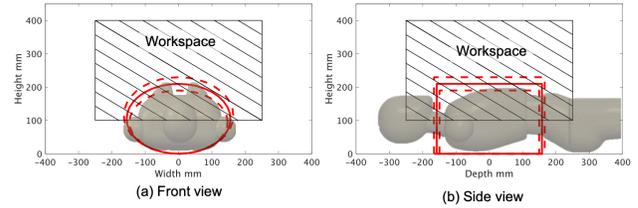

**Figure 3.** Workspace of the gantry-style positioning unit. Solid red line includes the mean of chest region size; dotted red line includes the standard deviation of chest region size and frame oblique lines demonstrates the reachable area of the US probe.

*B. Gantry-Style Positioning Unit Design*

The requirement of the gantry-style positioning unit is to satisfy the reachability and maneuverability to scan the whole chest with minimal risk of joint collision to the patient body.

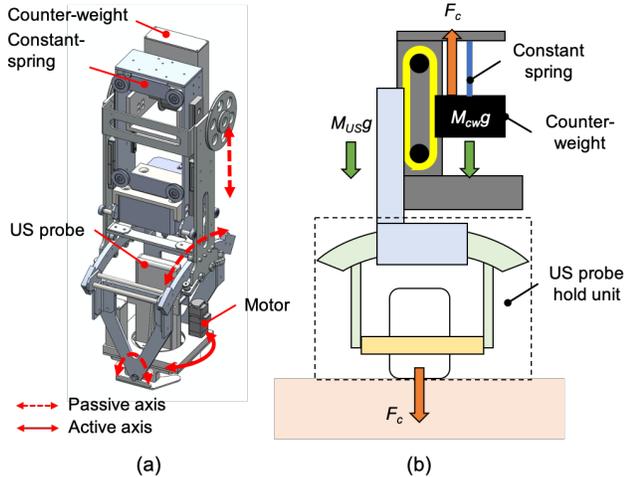
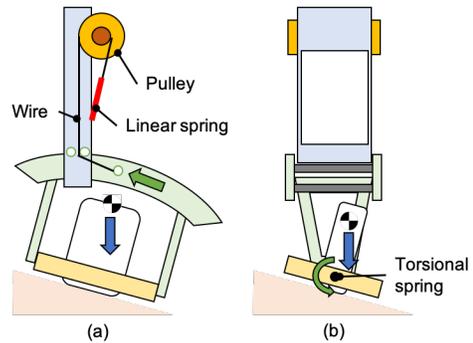

**Figure 4.** (a) Overview of the passive-actuated end-effector (red dot arrow: passive translation or rotation axis, red solid arrow: active rotation axis) (b) Passive mechanism of *z*-axis translation comprised with constant spring and counterweight.

The proposed positioning unit is comprised of 3 axes translation and 1 axis orientation. Referencing a human-body database created by the National Institute of Advanced Industrial Science and Technology [36], the representative chest lengths are as follows: depth 315.5 ± 16.5 mm; width 311.4 ± 17.1 mm; height 209.7 ± 19.9 mm; the required scan range can be defined approximately as an elliptical column with those lengths. To scan the lateral chest regions while maintaining the probe position perpendicularly to the surface, the orientation angle of end-effector needs to be rotated corresponding to the translation motion as shown in **Figure 2 (b)**. Based on the required scan range and the size of end-effector (see Sec. II-C), the linear actuators for 3-axis translations and the stepper motor for orientation are selected as shown in **Table I**. The workspace of the proposed configuration is shown in **Figure 3**, which satisfies the US probe reachability enough to scan the whole chest.

TABLE I. SPECIFICATION OF ACTUATORS

| Axis | Motion range (mm) | Product |
| --- | --- | --- |
| $x$ (transverse) | 1000 | BD07, Yamaha motor, Japan |
| $y$ (longitudinal) | 500 | SG07, Yamaha motor, Japan |
| $z$ (depth) | 350 | SG07, Yamaha motor, Japan |
| Rotation for end-effector | - | PKP564N28A2-TS30, Oriental motor, Japan |
| Rotation for probe | - | PKP214D06A, Oriental motor, Japan |

### C. Passive-Actuated End-Effector Design

The requirement of the end-effector is to hold the US probe while applying a certain force to the body surface. To acquire a diagnostic US image, an optimal tissue-probe contact force is required [23], [31]–[35]. In our previous work focusing on a robotic fetal ultrasonography [23], we proposed a passive mechanism to maintain the contact force via a constant spring and validated its feasibility through clinical studies. The advantage of the mechanical-based approach is to minimize the risk of electrical and computational failures (e.g. sensor failure and control error), and to absorb uncertain body motions and individual differences of the body shape. We extend optimized this concept for the LUS application. An overview of the proposed end-effector is shown in **Figure 4 (a)**. The key feature of the proposed end-effector is 1) to generate a constant force not dependent on the body size via a constant spring, 2) to maintain the US probe posture perpendicularly against the body surface with two passive rotational axes, and 3) to rotate the US probe angle for selecting the diagnostic view.

**Figure 5.** Passive mechanism to adjust the US probe position along (a) the craniocaudal direction and (b) the horizontal direction.

The configuration for maintaining the contact force within a certain range is shown in **Figure 4 (b)**. The force applied on the patient $F$ is formulated as:

$$F = F_C + (M_{CW} - M_{US}) g \quad (1)$$

where $F_C$ is the force generated by the constant force spring. $M_{CW}$ and $M_{US}$ are weights of the counterweight and US probe holding unit, respectively. We are able to maintain the force applied on the patient $F$ to be the same as the force generated by the constant force spring when $M_{CW} = M_{US}$. In the proposed system, we will initially use a constant force spring of 0.8 kgf, generating sufficient contact force while not providing subject discomfort according to the preliminary study [23].

During the LUS procedure, the US probe is required to maintain the normal direction with respect to the contact point on body surface. We designed a mechanism that permits adjustment of the US probe position passively along the craniocaudal direction (*y*-axis) and the horizontal direction (*x*-axis) (**Figure 5**). Along the craniocaudal direction, the passive mechanism that adjusts the US probe tip angle is realized by supporting the ring-shaped guide with a rotational shaft attached to a torsion spring. By implementing the torsion spring, the moment around the shaft due to the weight of the US probe is reduced and the position of the US probe is evenly perpendicular. To set the motion around the shaft to be zero, the following equation must be met:

$$m_p g l_p \sin\theta - k_p \theta = 0 \quad (2)$$

where $m_p$ is the weight of the US probe, $l_p$ is the distance between the center of rotation and center of gravity of the US probe. The $\theta$ is the position angle of the US probe along the craniocaudal direction, and $k_p$ is the torsion spring constant. Similarly, another passive mechanism is installed using a ring rail to enable the passive rotation along the horizontal direction.

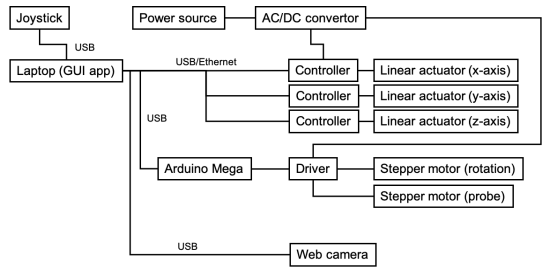

**Figure 6.** System architecture overview.

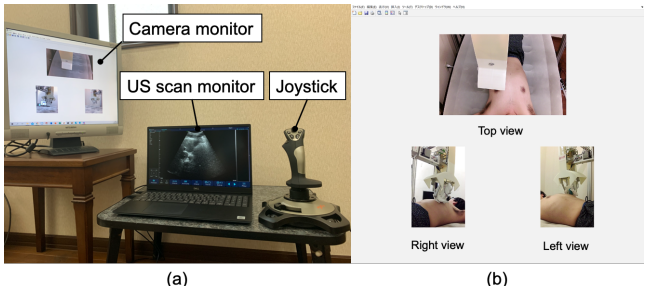

**Figure 7.** (a) Operational console overview and (b) web camera view to monitor the robot and interact with the patient.

The moment around the US probe tip can be expressed as follows:

$$m_r g l_r \sin\theta - k_r L^2 \sin(\varphi/2)\theta\sqrt{2 - 2\cos\theta} = 0 \quad (3)$$

where $m_r$, $l_r$, $k_r$, and $L$ represent the weight of the component including the passive mechanism and the US probe, the distance between the center rotation and center of gravity of the component, the linear spring constant, and the radius of the ring rail, respectively. The $\varphi$ is the angle of the US probe along the horizontal direction.

In addition to the 3 axes passive mechanism, 1 translation axis and 2 rotational axes were integrated as an active control to the remaining rotational axis along the *z*-axis; the axial direction with respect to the US image with a stepper motor (PKP214D06A, Oriental motor, Japan), permitting fine-tuning of the US image visualization based on the US probe angle and slice selection. For diagnosing lung disease, physicians often observe two orthogonal views (parallel and perpendicular to intercostal spaces) at each standardized scan region.

### D. Robot Control and System Architecture

The system architecture is summarized in **Figure 6**. The three linear actuators for translational motion in the positioning unit are controlled with a positioned type controller (TS-S2, Yamaha motor, Japan), which can generate pre-programable motions to the actuators. This occurs by receiving corresponding commands via serial communication, and the stepper motors for both orientations of end-effector and probe are controlled with an associated motor driver (CVD528B, Oriental motor, Japan) via a sequential pulse generated by an universal micro controller (Arduino Mega, Arduino, Italy). As for the interface for the robot operation, a joy-stick gaming controller (PXN-2113-SE, PXN, Japan) equipped with four-axis control and twelve programable buttons was utilized in terms of the simple and intuitive operation for physicians. We implemented two modes for the robot operation. First mode, called each axis control mode, is to control *x*-, *y*-, *z*-axis translations and end-effector and probe orientations individually. Tilting the joy-stick controller right-and-left, and back-and-forward are assigned to the *x*- and *y*-axis motions, respectively. Other axes motions are assigned to the programable buttons. Second mode, called arc motion mode, is to generate the arc motion around the body axis for moving the US probe from the anterior to side region effectively. In this mode, the *y*- and *z*-axis translations and the end-effector orientation are performed simultaneously when tilting the joy-stick controller only right-and-left. In short, each axis control mode is utilized when searching the diagnosable US images in each anterior, side, and posterior region, while the arc motion mode is utilized when shifting the scan region. The platform to integrate the interface and controllers was created in Matlab (Matlab, Mathworks, USA). An overview of the control console is shown in **Figure 7**. In addition, we used a wireless portable convex US probe (UProbe-C5, Sonostar, China). The US image visualization and acquisition control with the US probe can be conducted via an attached software remotely.

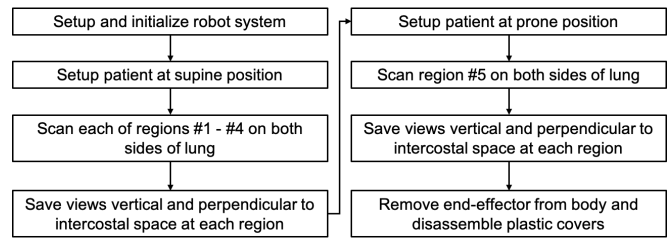

**Figure 8.** Workflow of the robot-assisted LUS scan.

### E. Workflow

The workflow of the robotic LUS operation is summarized in **Figure 8**. Initially, the robotic platform is transported to the location of the patient. The US probe is sterilized and covered with a plastic cover and is connected to the workstation. The operator sets the sterilized US probe at the end-effector and locates the end-effector to an initial position where it does not interfere with transporting the patient to the bed. These operations are completed while the patient is being prepared for the procedure outside the US imaging station. After the preparation, the patient is moved to the imaging station and placed on the bed in the supine position. The patient applies US gel on the chest as directed. During the scan phase, the operator moves the US probe to each standardized region as indicated in the modified BLUE protocol, collecting sequential US images until completion of the protocol. **Figure 1** shows the ten (5 x 2 planes) regions examined using the modified BLUE protocol. First, the end-effector holding the US probe is moved to region #1 of right side without contact with the patient. The US probe is placed on the chest and then moved downward until the passive mechanism along *z*-axis direction is activated. The operator searches for the diagnostic US image that are characteristic of normal or abnormal LUS features such as pleural line, A-lines, B-lines or lung sliding, by adjusting the end-effector position. The angle of the US probe was fixed parallel to the y-axis during the scan. Once the LUS features are observed, the operator adjusts the angle of the probe to acquire images perpendicular and parallel to the intercostal space and records each of the views for a period of 5 seconds. Next, the procedure is repeated in region #2. After scanning regions #1 and #2, the operator changes the control mode from each axis control mode to the arc motion mode and moves the US probe to the side of the chest following the arc path. The US probe position and posture are adjusted by changing the control mode to each axis control mode. Then, the image acquisition in regions #3 and #4 are repeated. Once the scans in regions #1 through #4 are completed for the right lung, an identical protocol is repeated for the left lung. Once the supine images are completed, the patient is placed in the prone position and region #5 imaging of both right and left lungs are scanned as described in **Figure 8**. Finally, the end-effector is removed from the body, and the plastic cover sterilizing the US probe is disassembled and discarded.

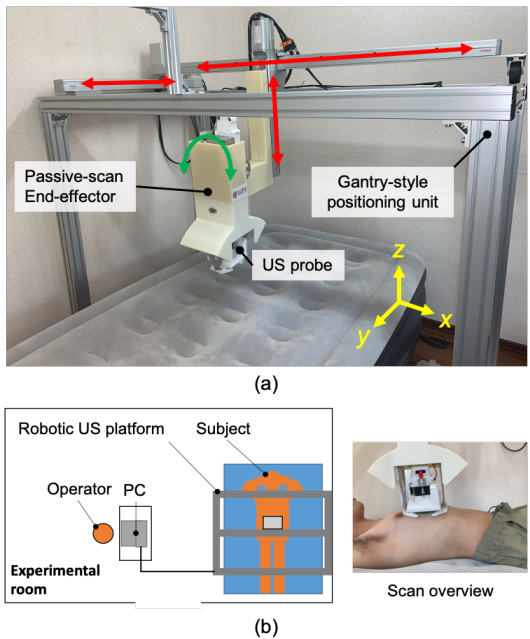

**Figure 9.** (a) Overview of the assembled LUS platform (b) Experimental setup overview.

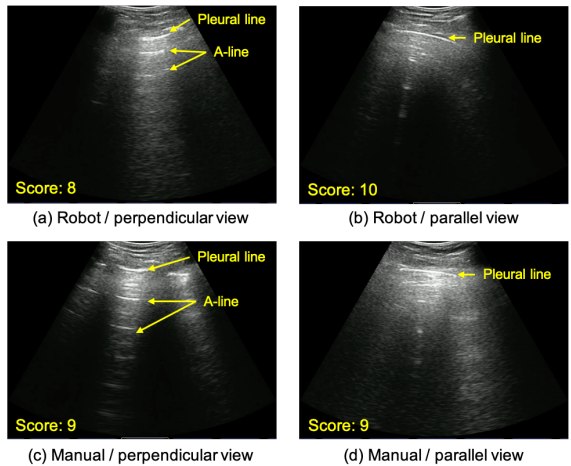

**Figure 10.** Examples of acquired images with the robot-assisted US and manual operator. Note the expected US reflectors were clearly identified by both methods.

*F. Experimental Setup in Humans*

**Figure 9 (a)** shows the overview of the assembled robotic LUS platform. For validating the feasibility of the proposed robotic LUS platform for further human subject studies and clinical trials, 1) the operational safety and 2) acquired image quality were quantitively assessed. For evaluating the operational safety quantitively, the contact force applied from the US probe during scanning each of the regions was measured with a 6-axis force/torque sensor (Nano 17, ATI Industrial, USA). The perpendicular force to the US probe tip was measured for 5 seconds in each of the 10 scan regions under the two subject's condition for evaluating the effects of respiration during the scan: allowing subjects to be imaged during normal respiration, and during breath hold. As the reference ground truth, the contact force during the manual operation was measured under the condition to allow subjects normal respiration. Two-tailed t-test with 90% confidence intervals were used to test the non-inferiority of the force data taken from the robot-assisted operation to that of the manual operation. We selected the non-inferiority margin of 2 N which is 10% of the acceptable contact force. Also, a sequential contact force that occurs during sliding the end-effector on the surface of chest was measured; the US probe was moved from region #1 to #2. For safety, a Visual Analog Scale (VAS) was used to monitor pain, feedback from the subjects. The level of discomfort experienced by the subjects was measured based on an established 11-point VAS, from 0 to 10. Due to the importance of comfort and safety of the subjects, a VAS > 4 (i.e., "hurts a little") was used as the point of termination.

To quantify the acquired image quality, the acquired images were scored by 2 experts in the field of LUS on COVID-19 patients and collaborators on this project (J.H., B.H.). The score of the image quality ranged from 0 to 10, with 0 indicating non-diagnostic quality and 10 for diagnostic superiority. The minimum score for diagnosis quality was set at 3. Criterion of the image acquisition is to observe the pleural line, A-line, and lung sliding. **Figure 10** shows the representative US image scored by the expert physicians. Additionally, the contrast-to-noise ratio (CNR) of pleural line to its surrounding area was calculated. The CNR is defined as:

$$CNR = \frac{|\mu_p - \mu_b|}{\sqrt{\sigma_p^2 + \sigma_b^2}} \quad (4)$$

where $\mu_i$ and $\mu_o$ are the means of the pixel value of the region of pleural line and background, respectively, and $\sigma_i$ and $\sigma_o$ represent the standard deviation of them. **Figure 9 (b)** shows the experimental setup overview. The bed size and its configurations are based on those commonly used in the hospital setting (height: 75 cm). The procedure followed the workflow outlined in Sec. II-E. As the ground truth, those images were acquired by the conventional manual operator following the same workflow and scored with the same criterion. In addition, the total durations of the procedure time were compared. To eliminate the bias of score depending on the physician's knowledge and perconception, the US images acquired by both robot and manual operations were assigned randomly and blindly to the readers. The non-inferiorities of the CNR, expert's score and total duration with the robot-assisted operation compared to those with the manual operation were tested by a two-tailed t-test with 90% confidence intervals. In the non-inferiority test of the CNR, expert's score and total duration, we selected those non-inferiority margins of 0.5, 2 point and 5 min, respectively, which were considered to be clinically significant.

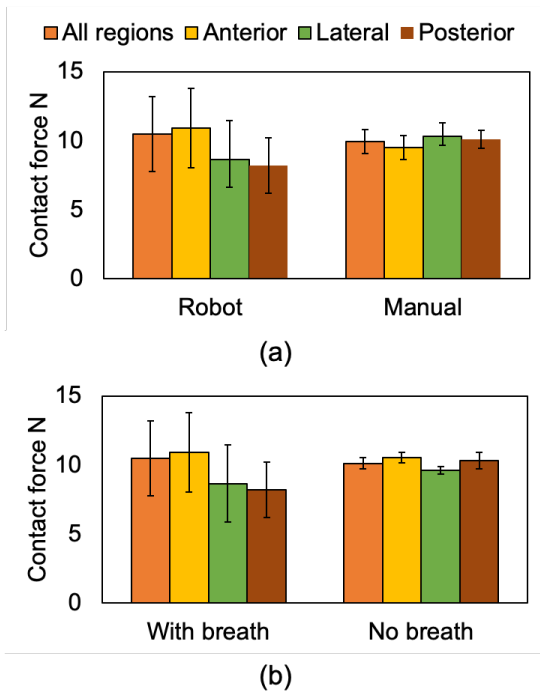

**Figure 11.** Mean contact force during the scan on each of the ten regions (a) comparing the robot-assisted and manual operations under the normal respiration and (b) in the robot-assisted operation comparing under the normal respiration and breath hold. The definition of region is shown in Figure 8.

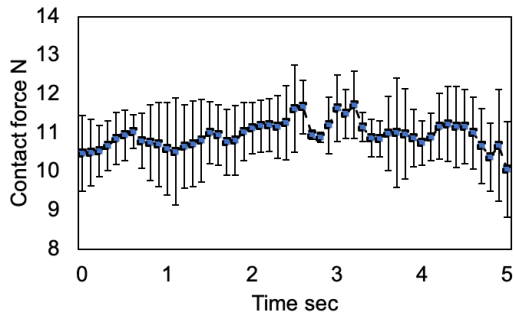

**Figure 12.** Time-series sequential contact force during the scan on the chest surface from region #1 to #2.

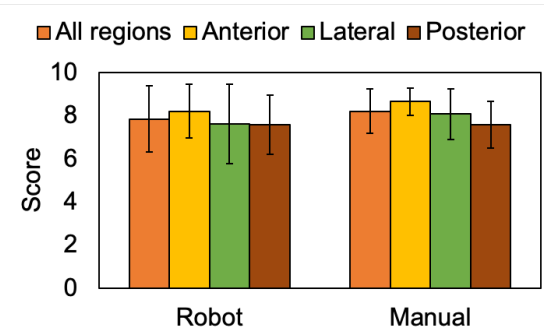

**Figure 13.** Mean expert reader scores of the US image quality acquired by the robot-assisted and manual operations. The definition of region is shown in Figure 8.

### G. Human Subjects and IRB Approval

The study was approved by the institutional research ethics committee at the Worcester Polytechnic Institute (No. IRB-21-

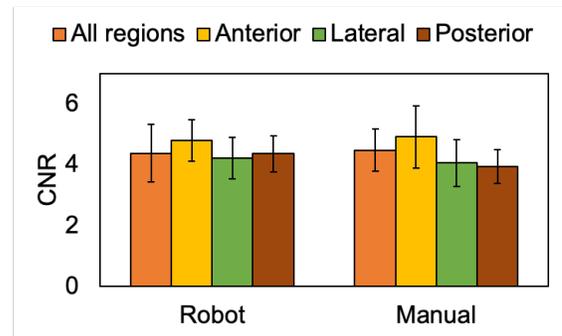

**Figure 14.** Mean CNR of pleural line captured in the US image acquired by the robot-assisted and manual operations. The definition of region is shown in Figure 8.

007), and written informed consent was given by the subjects prior to all test sessions. Three subjects were enrolled in the study (see **Table II** below).

TABLE II. SUBJECT INFORMATION

|  | Subject 1 | Subject 2 | Subject 3 |
|---|---|---|---|
| Age (yr) | 30 | 25 | 58 |
| Sex (M/F) | M | M | M |
| Height (cm) | 168 | 162 | 164 |
| Weight (kg) | 67 | 55 | 63 |
| Chest depth (cm) | 318 | 305 | 315 |
| Chest width (cm) | 315 | 298 | 308 |
| Chest height (cm) | 205 | 195 | 203 |

### III. RESULT

#### A. Operational Safety

The aim of this experiment is to validate the operational safety by evaluating the contact force between the US probe and body surface. We assumed that the contact force of 20 N was acceptable according to the previous studies [23], [31]–[35]. **Figure 11** shows the mean contact force during the scan on each of the region. The mean contact force in all regions measured by the manual and robot-assisted operations under allowing the subjects to breath normally was $9.52 \pm 1.02$ N and $10.48 \pm 2.72$ N, respectively. These results suggest there was no significant contact force difference in robot-assisted and manual operations ($p = 0.003$). Also, the mean contact force in all regions measured by the robot-assisted operation during breath hold was $10.13 \pm 0.38$ N. While there was no significant difference between each of those conditions, the standard deviation in the condition of robot-assisted operation under allowing the subjects to breath was large compared to other conditions. **Figure 12** show the sequential contact force during the scan on the chest surface from region #1 to #2. While the contact force during the sequential scan was fluctuated corresponding to the scan position compared to that during the statical scan, that was maintained within a certain range. Note that there was no report of greater than 4 points on the VAS in all patients.

#### B. Image Quality

The aim of this experiment was to compare and validate the image quality acquired by the robot-assisted and manual operations by scoring the acquired image by physicians and measuring CNR of the pleural line that is the primary feature of LUS. **Figure 13** shows the mean scores of the acquired image quality with robot-assisted and manual operations. The mean expert's score of all scan regions in the robot-assisted and manual operations were 7.85 ± 1.54 and 8.21 ± 1.04, respectively. Those results suggest the non-inferiority of the robot-assisted operation compared to the manual operation in terms of the visibility of lung features ($p = 0.004$). The acquired images with the robot-assisted operation successfully captured the pleural line and A-line which are representative signs for healthy lung as shown in **Figure 10 (a)**. Also, **Figure 14** shows the result of CNR of the pleural line captured in the acquired images. The mean CNR in the robot-assisted and manual operations were 4.38 ± 0.95 and 4.48 ± 0.70, respectively, showing no inferiority of the robot-assisted operation compared to the manual operation ($p = 0.005$). As expected, there was a statistical inferiority of the total procedure time of the robot-assisted operation compared to the manual operation (27.5 ± 5.4 min vs. 18.2 ± 3.2 min; ($p = 1$)).

*C. Discussion*

The results of this feasibility study demonstrated the safety, efficacy, and cost effectiveness of our tele-operative robotic LUS scan platform. The proposed platform enables a successful acquisition of LUS diagnostic images with an acceptable contact force in all the scan regions suggested in the BLUE protocol. The proposed robotic platform is capable of scanning all anterior, lateral, and posterior regions while maintaining the contact force within a safety range. The contact force was around 10 N (mean) and never exceeded 15 N, which ensures patient safety considering the safety limit of contact force reported in literature [23], [32]–[35] ranged from 20 to 24 N. No pain or discomfort was reported from the subjects during the scan in this study despite the small number of subjects. The applied contact force, on the other hand, is sufficient to maintain proper probe-skin contact for quality imaging of the diagnostic features of LUS. Both experts' score and CNR of all scan regions in the robot-assisted operation were not inferior to those of manual operation.

Regarding the remote operation, the number, location and angle of the camera may need further optimization, we located three cameras to fully cover the top and both side of chest and successfully captured US images at all scan regions required in the modified BLUE protocol. Without investigating the mental/cognitive workload for the remote operation with respect to the configuration of cameras, the procedure duration in the robot-assisted operation was significantly longer than in the manual operation. The limited visibility due to the current configuration of cameras may cause increased procedure time. However, we believe the additional time required by the robotic platform is a reasonable trade-off for the safety and mitigation in the spread of disease in those patients who are potentially highly infectious. It is necessary to optimize camera configuration and user-interface to reduce the procedure duration in this robot-assisted operation. The user-interface can also be improved for remote US diagnosis. While we utilized a general joystick as the interface, previous tele-operative US robot systems utilized several types of user-interface such as graphic user interface (GUI) [27], master manipulator [14] and joystick-like interface incorporating haptic feedback [37], [38]. Also, for assisting remote operations effectively, incorporating an augmented reality (AR)-based navigation system such as [39] into the tele-operative US scan may contribute to reduce the procedure time.

A major limitation of the proposed platform is that we cannot perform fine adjustment of US probe alignment and contact force actively. The proposed end-effector orients the US probe normal to the body surface independent from the scan positions. While the pleural line was captured by the robot-assisted operation in all scan regions, some features such as A-line were not always visible due to the lack of rotation adjustment. Given the manual operator needs to adjust the US probe orientation finely to obtain such the feature clearly, additional actuators for rotation may be implemented into the end-effector. Furthermore, when acquiring the US images in patients with excessive subcutaneous tissue, the manual operator often applies addition force to the US probe. Consequently, the contact force applied by the end-effector need adjustments based on the patient's body habitus.

Another significant limitation of this study is the lack of validation and/or cross-validation of other imaging modalities in COVID-19 patients. Several pathological features including the disappearance of lung sliding, B-line increase, and lung consolidation cannot be evaluated in the healthy subjects. Before implementing the proposed platform for COVID-19, it is necessary to further evaluate the platform on a larger number subjects including those with non-COVID-19 related lung disease. Additional improvements in the hardware design must be taken into account. Because the procedure to apply US gel to the thorax is still manual, this confers a potential risk of transmission of infection. To perform fully tele-operative LUS, an additional autonomous gel-dispenser should be incorporated with the end-effector.

IV. CONCLUSION

This manuscript presents a feasibility study of applying a tele-operative low-cost robotic platform on human subjects for the application of LUS. Our results demonstrate that the proposed platform enables successful acquisition of diagnostic images for LUS with a safe contact force at all standardized scan regions using the modified BLUE protocol on healthy subjects. There were no significant differences between the measured contact force, the CNR of pleural line, and image quality scored by physicians between the two methods. While the total procedure duration in the robot-assisted operation is longer than that in the manual operation, we believe the additional time required by the robotic platform is a reasonable trade-off for the safety and mitigation in the spread of disease in those patients who are potentially highly infectious. We are optimistic that the total procedure duration can be reduced by optimizing the camera configuration and incorporating a navigation system within the proposed platform. The robotic LUS platform presented in this study has the potential to be applied to COVID-19 and other highly infectious diseases. The use of this technology may minimize risk of disease transmission among patients and healthcare workers in a resource-limited environment by minimizing the physical contact during the US procedure.

ACKNOWLEDGMENT

Financial support was provided in part by the National Institute of Health (DP5 OD028162). The authors would like to thank Yamaha Motor, co. ltd., Ono-denki, co., Sonostar, co., Dr. Y. Uchida, and Mr. JT. Kaminski for assisting this project.